\documentclass[10pt,conference,a4paper]{IEEEtran}

\IEEEoverridecommandlockouts
\usepackage{cite}
\usepackage{amsmath,amssymb,amsfonts}
\usepackage{algorithmic}
\usepackage{graphicx}
\usepackage{textcomp}
\usepackage{tablefootnote}
\usepackage[flushleft]{threeparttable}
\usepackage{marvosym}
\usepackage{xcolor}

\def\BibTeX{{\rm B\kern-.05em{\sc i\kern-.025em b}\kern-.08em
    T\kern-.1667em\lower.7ex\hbox{E}\kern-.125emX}}
    
\usepackage{diagbox}
\usepackage{booktabs}
\usepackage{multirow}
\usepackage{amstext}
\usepackage[numbers,sort&compress]{natbib}
\usepackage{subfig}
\usepackage{float}
\usepackage[colorlinks,
            linkcolor=blue,
            anchorcolor=black,
            citecolor=green]{hyperref}
\usepackage{url}
\begin{document}
\title{Dynamically Mitigating Data Discrepancy with Balanced Focal Loss for Replay Attack Detection
%{\footnotesize \textsuperscript{*}Note: Sub-titles are not captured in Xplore and
%should not be used}
%\thanks{This research work was supported by the Fundamental Research Funds for the Central Universities (Grant No. 2016JX06); and the World-Class Discipline Construction and Characteristic Development Guidance Funds for Beijing Forestry University (Grant No. 2019XKJS0310).}
}
\author{\IEEEauthorblockN{Yongqiang Dou\textsuperscript{1 3},
Haocheng Yang\textsuperscript{1 3}, Maolin Yang\textsuperscript{1 3}, Yanyan Xu\textsuperscript{1 3}\textcolor{black}{\textsuperscript{ \Letter}}, Dengfeng Ke\textsuperscript{2}\thanks{\textsuperscript{ \Letter}
% \IEEEauthorrefmark{5}
Corresponding author.}$\footnote{Corresponding author}$
\textsuperscript{\Letter}% \IEEEauthorrefmark{5}
\\\\}
\IEEEauthorblockA{\textsuperscript{1}School of Information Science and Technology,
Beijing Forestry University, 
Beijing 100083, China\\ 
}
\IEEEauthorblockA{\textsuperscript{2}National Laboratory of Pattern Recognition, 
Institute of Automation, Chinese Academy of Sciences, 
Beijing 100190, China}
\IEEEauthorblockA{\textsuperscript{3}{\small Engineering Research Center for Forestry-oriented Intelligent Information Processing, National 
Forestry and Grassland Administration}\\ 
\textsuperscript{\Letter} xuyanyan@bjfu.edu.cn, dengfeng.ke@nlpr.ia.ac.cn}

}

%
%\author{\IEEEauthorblockN{1\textsuperscript{st} Yongqiang Dou }
%\IEEEauthorblockA{\textit{School of Information Science and Technology} \\
%\textit{Beijing Forestry University}\\
%Beijing 100083, China \\
%YongqiangDou@ieee.org}
%\and
%\IEEEauthorblockN{2\textsuperscript{nd} Haocheng Yang}
%\IEEEauthorblockA{\textit{School of Information Science and Technology} \\
%\textit{Beijing Forestry University}\\ Beijing 100083, China \\
%yhc2903107062@bjfu.edu.cn}
%\and
%\IEEEauthorblockN{3\textsuperscript{rd} Maolin Yang}
%\IEEEauthorblockA{\textit{School of Information Science and Technology} \\
%\textit{Beijing Forestry University}\\
%Beijing 100083, China\\
%yangmaolin@bjfu.edu.cn}
%\and
%\IEEEauthorblockN{4\textsuperscript{th} Yanyan Xu}
%\IEEEauthorblockA{\textit{School of Information Science and Technology} \\
%\textit{Beijing Forestry University}\\
%Beijing 100083, China \\
%xuyanyan@bjfu.edu.cn}
%\and
%\IEEEauthorblockN{5\textsuperscript{th} Given Name Surname}
%\IEEEauthorblockA{\textit{National Laboratory of Pattern Recognition}\\
%\textit{Institute of Automation, Chinese Academy of Sciences}\\
%Beijing 100190, China \\
%dengfeng.ke@nlpr.ia.ac.cn}
%
%}

\maketitle

\begin{abstract}
It becomes urgent to design effective anti-spoofing algorithms for vulnerable automatic speaker verification systems due to the advancement of high-quality playback devices. Current studies mainly treat anti-spoofing as a binary classification problem between bonafide and spoofed utterances, while lack of indistinguishable samples makes it difficult to train a robust spoofing detector. In this paper, we argue that for anti-spoofing, it needs more attention for indistinguishable samples over easily-classified ones in the modeling process, to make correct discrimination a top priority. Therefore, to mitigate the data discrepancy between training and inference, we propose D3M, to leverage a balanced focal loss function as the training objective to dynamically scale the loss based on the traits of the sample itself. Besides, in the experiments, we select three kinds of features that contain both magnitude-based and phase-based information to form complementary and informative features. 
Experimental results on the ASVspoof2019 dataset demonstrate the superiority of the proposed methods by comparison between our systems and top-performing ones. Systems trained with the balanced focal loss perform significantly better than conventional cross-entropy loss. With complementary features, our fusion system with only three kinds of features outperforms other systems containing five or more complex single models by 22.5\% for min-tDCF and 7\% for EER, achieving a min-tDCF and an EER of 0.0124 and 0.55\% respectively. Furthermore, we present and discuss the evaluation results on real replay data apart from the simulated ASVspoof2019 data, indicating that research for anti-spoofing still has a long way to go.  Source code, analysis data, and other details are publicly available at 
\url{https://github.com/asvspoof/D3M}.
%老师就是这个链接之前说放实验部分，但是我看有的顶会文章也是放第一页，一眼开源会不会比较加分
%Diverse feature representations and 

\end{abstract}


\section{Introduction}
\begin{figure}
\centering
\includegraphics[width=\columnwidth]{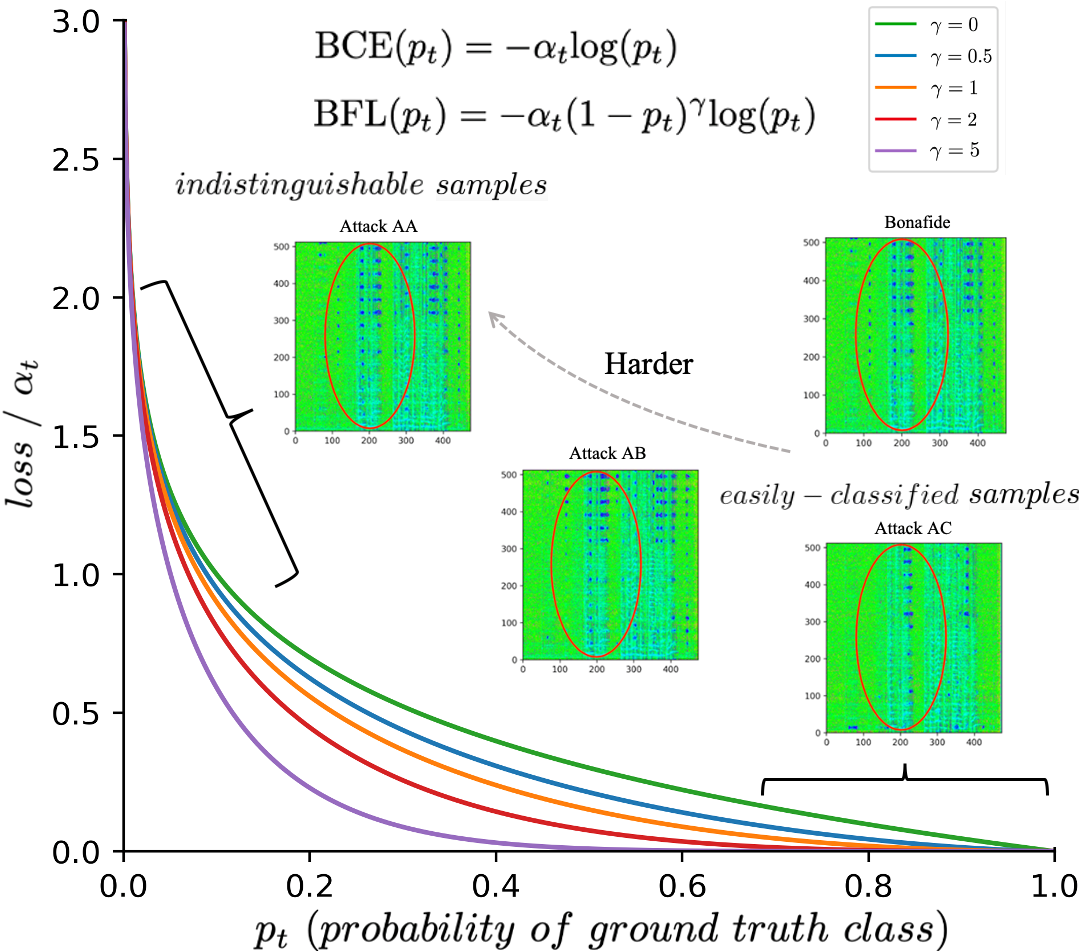}
\hspace{8em}
\caption{An example for comparison between a bonafide utterance and its corresponding spoofed utterances with different attack types, illustrates that high-quality attack AA has only subtle differences from the bonafide one, yet taking up a small portion of the data. We term this phenomenon as \textit{data discrepancy} in anti-spoofing, and propose a method to replace the conventional {balanced cross-entropy loss (BCE)} with {the novel balanced focal loss (BFL)} as {the training} objective. 
It is worth noting that hard and easy samples do not have strict boundaries, nor will there be a certain attack method as the dividing line, which demonstrates the necessity of making dynamic adjustments with BFL.
 The MGD-gram feature is used for visualization, see Fig.\ref{fig:sali} for high-resolution visualization. Description of different attack types (AA, AB, AC) and analysis are elaborated in Section \ref{sec:settings} and Fig. \ref{fig:sali}, respectively. Best viewed in color. }
\label{fig:main}
\end{figure}

Automatic Speaker Verification (ASV), intended for authenticating a claimed speaker identity by characteristics of the voice, has shown promising results recently. With rapid advancement of {ASV and} its wide applications such as smart assistants and banking systems, vulnerability of ASV systems has been gradually exposed. More specifically, current ASV systems have almost no defense against spoofing attacks (also known as presentation attacks). According to possible attack locations in a typical ASV system (ISO/IEC 30107$\footnote{https://www.iso.org/standard/67381.html}$), spoofing attacks can be categorized into four major classes: (i) impersonation, (ii) speech synthesis, (iii) voice conversion and (iv) replay. The first three classes rely on professional knowledge heavily, while replay attack does not require any kind of expertise. Besides, 
easy access to high-quality playback devices makes it even more urgent to develop robust anti-spoofing systems against replay attack, which is also the goal we strive for in this work.

Thanks to the impetus injected by the 2015, 2017 and 2019 ASVspoof Challenges, great achievements have been made by numerous researchers in the past few years. In 2017, replay attack detection was first introduced to the challenge with the aim of {measuring limits} and developing countermeasures. Simultaneously, deep learning-based approaches came to the fore \citep{LCNN2017}. The ASVspoof2019 Challenge extended the previous challenge with improved, controlled simulation and start-of-the-art spoofing methods for generating replay data, as well as new primary evaluation metric t-DCF \citep{Kinnunen2018}. Such anti-spoofing systems as \cite{LCNN2019} and \cite{ASSERT2019} ranked among the best systems, leading to two lines of work for deep learning-based architectures. One is based on LightCNN, and the other is based on ResNet \citep{Challenge2019results}.

Current state-of-the-art anti-spoofing systems mainly suffer from two challenges. On the one hand, there exists discrepancy in data distribution among training, testing, evaluation and real {data, which} has a great impact on the model performance. Similar to image classification, previous {studies} like \citep{Alzantot2019,DKU2019} regard the data discrepancy as a class-imbalance problem. Widely-adopted strategies to solve the class-imbalance problem can be divided into re-sampling and re-weighting \cite{Cui2019}. Re-sampling strategies, including over-sampling and under-sampling, are complicated and have drawbacks such as incurring risks from removing important samples.
Compared with re-sampling, re-weighting strategies are relatively simple and based on the statistics of data, for example, use the inverse of class frequency as the weighting factor \cite{Alzantot2019}.
For replay attack detection, we argue that, the main challenge lies in the data discrepancy and it cannot only be viewed as class-imbalance problems of binary classification\cite{Alzantot2019} or multi-class classification \citep{ASSERT2019} in a narrow sense. Strategies for the imbalance of different classes place more emphasis on the inter-class fairness, 
while for data discrepancy, more emphases are needed for easily-misclassified indistinguishable samples, which only take up a small portion in the training data, but	 are the overriding factors as the attack sources come from increasingly accessible quality devices\cite{Focalloss,Cui2019}. Therefore, a dynamically re-weighting training objective becomes a crucial driving force to bridge the gap.
On the other hand,  the need to select informative feature representations arises when building a system with a growing number of features. As a consequence, choosing fewer but sufficient, complementary features is of supreme importance \cite{featuredecision}.
In this paper, we focus on resolving the two challenges mentioned above, {especially the} first one. Our main contributions can be summarized as follows:
\begin{enumerate}

\item Inspired by \cite{Focalloss}, we  propose a \textbf{D}ynamic \textbf{D}ata \textbf{D}iscrepancy \textbf{M}itigation method (D3M) that leverages balanced focal loss as a novel training objective for anti-spoofing, which enables the model to attend more to indistinguishable samples with dynamically scaled loss value. Through detailed analysis, we find that balanced focal loss outperforms other baselines to a large extent. To our knowledge, we are the first to introduce focal loss to anti-spoofing, mitigating the data discrepancy between
training and inference.

\item Based on our survey of the ASVspoof2019 Challenge, only group delay (GD) was used by researchers as phased-based feature{s}. We extend the ideas from \cite{Tom2018} to first investigate the performance of the modified group delay function \citep{mgd2004}, dubbed MGD-gram, on the improved ASVspoof2019 dataset which uses start-of-the-art spoofing methods for generating replay data. Also, we demonstrate the superiority of fusion of three kinds of complementary features, namely modified group delay (MGD) gram, short-time Fourier transform (STFT) gram and constant Q transform (CQT) gram.

\item We show that the performance of current top-performing systems on real data are not as good as on the simulated ASVspoof2019 data \citep{DATASET}, which is unexpected and considered  very  worthy  of  discussion. This may be due to the fact that simulated data cannot be applied to real cases, or the distinctions between GMM and ConvNets. Deep learning-based methods for anti-spoofing still {have} a long way to go, as the conventional GMM model has the best performance, although it is not good enough, with an EER of 12.4\%.

\end{enumerate}
\section{Cost-Sensitive Training --- the Balanced Focal Loss}

Almost all the anti-spoofing systems have 
poor performance for samples made by replay devices with higher quality and a shorter attacker-to-talker distance. These indistinguishable samples can be easily misclassified, thus posing a severe threat to anti-spoofing systems. Moreover, indistinguishable samples only take up a small portion of the training data, {making the recognition} extremely harder because the training procedure is dominated by the majority. Specifically, for gradient-based methods like neural networks, the gradients are dominated by easy samples. 

We term the above phenomenon as the data discrepancy between training and inference in anti-spoofing. 
Unlike image classification or object detection suffering from {class-imbalance,} data discrepancy faced in anti-spoofing is a more severe challenge, which emphasizes more on the correctness of discrimination and the security of the biometric system than just on the fairness between different classes of attacks. 

To mitigate this problem, on the basis of the common strategy to solve class imbalance, i.e. simple re-weighting (balancing by class frequency), we further propose to {leverage BFL as the training objective instead of BCE.} Focal loss was first used in the field of object detection, and its validity has been tested on many tasks \citep{Focalloss}. 

We use the example shown in Fig. \ref{fig:main} to illustrate our idea.
%We use the example in Figure \ref{fig:comparison} to illustrate our idea.
As shown in Fig. \ref{fig:main}, only subtle differences {exist} with red marks for the sample of attack type AA, compared with other two samples of ``easier'' attack types such as AB and AC. %这个地方没有同一句话，不太现实，需要修正

Intuitively, to increase the accuracy for harder samples of attack type AA, we need to make the system pay more attention to them during training. Therefore, simply by assigning bigger loss values to harder samples and smaller values to easier {examples, we} can achieve our goal. 

Formally, the \textit{balanced focal loss}, a weighted variant of the standard focal loss, can be calculated as:
\begin{equation}
\mathrm{BFL}\left(p_{\mathrm{t}}\right)=-\alpha_{\mathrm{t}}\left(1-p_{\mathrm{t}}\right)^{\gamma} \log \left(p_{\mathrm{t}}\right),
\end{equation}
where subscript $\mathrm{t}$ refers to the true class label,
and $\alpha$ denotes the weight for the corresponding class to mitigate the class-imbalance problem, and $\gamma$ is tunable as a focusing parameter to control the relative {scaling.  We} find $\gamma=2$ best in the experiments {(the red curve} in the Fig. \ref{fig:main}).

% \begin{figure}[h]
% \centering
% \includegraphics[width=0.95\columnwidth]{images/loss_illu.jpg} 
% \caption{Illustration of training objective BFL and BCE Loss used in this work. Both of them uses $\alpha_t$ (the inverse of class frequency) to statically re-weighting the loss, whilst BFL leverages additional $\left(1-p_{\mathrm{t}}\right)^{\gamma}$ to dynamically scale the loss value based on the probability of target label $p_{\mathrm{t}}$ and a focusing factor $\gamma$. $\gamma$ is tunable to control the relative scaling and we find $\gamma=2$ best in the experiments (red curve in the figure).}
% \label{fig:focalillu}
% \end{figure}

{Both BFL and BCE use $\alpha_t$ (the inverse of class frequency) to statically re-weight} the loss. The main difference between BFL and BCE is that BFL uses an additional weighting factor $\left(1-p_{\mathrm{t}}\right)^{\gamma}$ to dynamically scale the value of the contribution of each sample to the final loss, based on the probability of target label $p_{\mathrm{t}}$ and a focusing factor $\gamma$, {so as to} focus more on
indistinguishable samples and reduce
the relative loss for easily-classified samples, as illustrated in Fig. \ref{fig:main}.
It is worth noting that since $p_{\mathrm{t}}$ is the probability of the target label predicted by the weight, {using $\left(1-p_{\mathrm{t}}\right)^{\gamma}$} to scale the loss is like performing a soft attention. 
%Focal loss was first used in the field of object detection \citep{Focalloss}, and its validity has been tested on many tasks. With gratifying results produced on those tasks, we believe it is suitable for replay attack detection against quality samples.

\section{The Model Architecture --- End-to-end Residual Network}
% \begin{figure*}[h]
% \centering
% \includegraphics[width=.95\textwidth]{images/model.png} 
% \caption{Illustration of our model architecture. Best viewed in color.}
% \label{fig:network}
% \end{figure*}

\begin{table}[h]
 \renewcommand\arraystretch{1.2}
 \centering

 \caption{The network structure used in this paper. The input feature size is (1, 513, 500). Any two convolutional layers have batch normalization followed by the ReLU activation function in between. Global average pooling is used between the last convolutional layer and the fully-connected layer.}

 \label{table:network} 
\begin{tabular}{c|l|l|l}
\toprule[1.5pt]
Layer & Filter & Output shape & \# Params.\\
\midrule[1pt]

 Conv2d     &3x3,1  &(16, 513, 500)& 144 \\
 Max Pooling & --  & (16, 513, 500) & --\\
ResBlock1 x 3 & 3x3,16    & (16, 513, 500) & 4.6k x 3 \\
ResBlock2 x 4       & 3x3, 32 & (32, 257, 250) & 18.4k x 4  \\
ResBlock3 x 6   & 3x3, 64 & (64, 129, 125) & 73.7k x 6 \\
ResBlock4 x 3 & 3x3, 128    & (128, 65, 63)  & 295.0k x 3 \\
 GAP         & --               & (128, ) & -- \\ \hline
 FC          & 32 & (32, )           & 4.1k     \\
 Output      & 2 & (2, )  & 66        \\

\bottomrule[1.5pt]

\end{tabular}
 \end{table}

Recently, with great progress made by researchers participating in the 2015, 2017 and 2019 ASVspoof challenges,  LightCNN-based and ResNet-based deep neural networks have become the mainstream as high-level feature extractors \citep{LCNN2017,LCNN2019,ASSERT2019}.
As mentioned earlier, instead of proposing a novel network architecture, this work mainly aims to present informative feature representations and the effective training objective. Hence, we select the ResNet-based end-to-end model as the backbone, making use of its superior extracting and modeling capabilities. 

As illustrated with detailed configurations in Table \ref{table:network}, our model is similar to \cite{DKU2019}. The main differences are: (i) fixed-length feature representations are used, and features are either padded or truncated to n\_frames = 500 along the time axis according to statistics derived from data, whereas in \cite{DKU2019}, the model takes fixed-length input utterances for training and variable-length utterances {for test}. The purpose of our modification behind is to facilitate the training process while maintaining consistency during training and inference; (ii) all the models are trained from scratch with no modification {of} data, that is, we  use neither data augmentation nor pre-trained techniques.  

The final countermeasure score, representing the genuineness judgement for each utterance provided by the system,  is calculated as the log-likelihood ratio using Eq. \eqref{eqn:lld}, 
\begin{equation}
\label{eqn:lld}
s_{cm}(utt)=\log (p(\text{bonafide} | utt ; \theta))-\log (p(\text{spoof} | utt ; \theta))
\end{equation}
where $utt$ refers to {a test} utterance, and $\theta$ denotes model parameters. This is also a recommended method for score computation by the ASVspoof2019 Commmittee. The probabilities of bonafide and spoofed speech utterances are given by the final softmax layer of the model.

%\subsection{Equations}
%Number equations consecutively. To make your 
%equations more compact, you may use the solidus (~/~), the exp function, or 
%appropriate exponents. Italicize Roman symbols for quantities and variables, 
%but not Greek symbols. Use a long dash rather than a hyphen for a minus 
%sign. Punctuate equations with commas or periods when they are part of a 
%sentence, as in:
%\begin{equation}
%a+b=\gamma\label{eq}
%\end{equation}
%
%Be sure that the 
%symbols in your equation have been defined before or immediately following 
%the equation. Use ``\eqref{eq}'', not ``Eq.~\eqref{eq}'' or ``equation \eqref{eq}'', except at 
%the beginning of a sentence: ``Equation \eqref{eq} is . . .''

\section{Integration of Complementary Features}
\label{sec:feature}
Magnitude-based information included in short-time Fourier transform gram has been widely used in top-performing anti-spoofing systems such as \cite{ASSERT2019} and \cite{Alzantot2019}, while features containing phased-based information used in \cite{Tom2018} and \cite{DKU2019} also yield superior performance. However, in the ASVSpoof2019 Challenge, we find that only GD-gram was used by researches as phased-based time-frequency representation, although Modified GD feature has been proved {effective} in the previous work \cite{mgdxiao2015a,mgdliuyi2015a}. To further investigate their performance on the improved ASVSpoof2019 dataset, we {compare MGD-gram} with GD-gram in out experiments. Both of them are low-level time-frequency representations, which can better utilize the modeling capabilities of the ResNet neural network.
Besides, to integrate complementary features, in this paper, we carefully explore three different kinds of features, namely MGD-gram, STFT-gram and CQT-gram. % 加一些选择这三个features的原因
We employ the MGD-gram and STFT-gram to integrate both magnitude and phase information. Also, the CQT-gram, shown in \cite{cqccpropose} to yield a superior performance to general forms of spoofing attack with {higher frequency resolution in the lower frequency(?)}, is added to them to form complementary informative feature representations.

\subsection{Modified Group Delay Gram (MGD-gram)}
In \cite{Tom2018,DKU2019}, the group delay function, defined as the negative derivative of phase, was used to {characterize speech} signals to distinguish bonafide utterances from spoofed ones on the ASVspoof2019 dataset. 
%is:
%\begin{equation}
%\tau(\omega, t)=-\frac{d(\theta(\omega, t))}{d \omega},
%\end{equation}
%where $\theta(\omega,t)$ is the unwrapped phase spectrum at frequency $\omega$ and time $t$. 
%It 
%can be directly computed from the speech signal using the following equation:
%\begin{equation}
%\tau(\omega, t)=\frac{X_{R}(\omega, t) Y_{R}(\omega, t)+Y_{I}(\omega, t) X_{I}(\omega, t)}{|X(\omega, t)|^{2}},
%\label{eqn:gd}
%\end{equation}
%where R and I denote the real and imaginary parts of the Fourier transform. $X(\omega, t)$ and $Y(\omega, t)$ are the Fourier transforms of $x(n)$ and $nx(n)$ in a frame $t$, respectively.
%这里说n(x)为了说明Y是什么
%The final GD-gram representation is formed by concatenating the outputs of Eq. \ref{eqn:gd} for all frames in an utterance.

As clearly illustrated in \cite{mgd2003}, the vanilla group delay function suffers from its spiky nature and requires the signal be a minimum phase. However, speech segments can be non-minimum due {to zeroes} from windowing and noise. 
%We therefore extend the  group delay function to a modified group delay function to form the novel MGD-gram feature representations. 
The MGD function is thus proposed as a parameterized and improved version of the GD function, which is formulated using  \cite{mgd2004}:
\begin{equation}
\tau_{m}(\omega,t)=\left(\frac{\tau(\omega,t)'}{|\tau(\omega,t)'|}\right)(|\tau(\omega,t)'|)^{\rho},
\end{equation}
where
\begin{equation}
\tau(\omega,t)'=\left(\frac{X_{R}(\omega,t) Y_{R}(\omega,t)+Y_{I}(\omega,t) X_{I}(\omega,t)}{S(\omega,t)^{2 \lambda}}\right),
\end{equation}
where $S(\omega,t)$ is the cepstrally smoothed version of $|X(\omega,t)|$, and $\rho$ and $\lambda$ are newly added parameters to reduce the aforementioned spikes ($0 < \rho \leq 1.0$ and $0 < \lambda \leq 1.0$).

After replacing the GD function with the MGD function, we can easily get the MGD-gram representation by concatenating all the frames' outputs. 
%Note that parameters $\rho$ and $\lambda$ need to be tuned (we find $\rho = 0.2$ and $\lambda = 0.7$ best in our experiments)

\subsection{Short-time Fourier Transform Gram (STFT-gram)}
The short-time Fourier transform (STFT) \cite{Zhao2007Voltage} was proposed to solve the problem that the Fourier transform cannot reflect local features of a signal. The utterances are first broken up into overlapping frames and then the Fourier transform is performed on each short frame, forming a 2-D complex matrix finally. {STFT  converts a time domain signal into a frequency} domain signal. {STFT-gram} (also know as spectrogram) is one of the most widely-used features now, which contains magnitude-based information. In \cite{ASSERT2019}, many systems achieved great performance {with STFT-gram.} As a result, we choose STFT-gram as one of our complementary features.

\subsection{Constant Q Transform Gram (CQT-gram)}
The constant Q transform (CQT) \cite{Brown1998Calculation} employs geometrically spaced frequency bins to make the constant Q factor across entire spectrum. {CQT} was designed to resolve problems for musical temperament as it can give the same frequency as the scale frequency. {Simultaneously}, CQT performs well {for} automatic speaker verification with a higher frequency resolution at lower frequencies and a higher temporal resolution at higher frequencies, which enables it closer to {human perception.} Thus, {we incorporate CQT-gram} into our complementary features. We apply CQT on the utterance, {and log transform} is followed to derive the CQT-gram.

\subsection{Fusion Scheme}
%Typical fusion methods include early fusion and late fusion. As the name implies, early fusion is to perform feature-level fusion, which concatenates different features, and input them into a model for training. Late fusion refers to the fusion of the prediction scores of different systems (score-level fusion). To be more specific, late fusion is to train several models, and then fuse all  prediction scores predicted by these models to get the final result. Common late fusion methods include simply taking average, maximum and weighted average of the scores, or using logistics regression.

In our experiments, we train three models
that have identical model architecture with aforementioned features, and later fuse their scores by taking the average (mean-fusion) and employing logistics regression (LR).
Note that we strictly follow the evaluation protocol from the ASVspoof2019 Challenge \citep{DATASET}. 
%The reason of using average value is that its calculation does not need complicated process and can get a satisfactory result in a short time. In the experiment, we found that sometimes a lower t-DCF value could be obtained by using average value, and sometimes a lower t-DCF value could be obtained by using logistics regression. Through careful study of the code, it was found that the effectiveness of using the logistic regression for late fusion depends on the number of iterations. If the number of iterations is too small, the obtained t-DCF value is even higher than the result obtained by averaging. On the contrary, if the number of iterations is too large, the training time is too long. By constantly adjusting the parameters, we can get a reasonable number of iterations and get a lower t-DCF value in a shorter training time. When the number of iterations is 200, the t-DCF value obtained by using logistics regression is 0.0125, while the t-DCF value obtained by using the average value is 0.0128.

\section{Experiments \& Analysis}
Experiments in this study were conducted using PyTorch \cite{NEURIPS2019_9015},
a deep learning library in Python. Source code and other details are publicly available at 
\url{https://github.com/asvspoof/D3M}.
\subsection{Baseline Systems}
\textbf{Offical Baselines:} We adopt the {officially two baseline models together with the dataset released} by the ASVspoof 2019 Committee. These two systems are based on the same conventional 2-class GMM backend with 512 components and two kinds of acoustic features, {namely linear} frequency cepstral coefficients (LFCC) and constant Q cepstral coefficients (CQCC). Details can be found in \cite{DATASET}.

\textbf{Top-performing NN-based Models: } According to the challenge results reported in \cite{Challenge2019results}, many top systems employ neural network(NN)-based models. To further test the performance of our proposed methods, {we choose} the newly published systems ranked 3rd sidual architectures to our NN-based baseline models \citep{ASSERT2019,DKU2019}{. The} top-1 system of this challenge has not been publicly available yet, and the top-2 system is a LightCNN based one with {some modifications} on feature sizes which is not very suitable for comparison. Specifically, we adopt the ResNet architecture trained with the balanced cross-entropy loss as our baseline models \citep{DKU2019}. Note that in order to control variables, we further {re-implemented} the work in \cite{DKU2019} under the same settings with a similar network architecture.

\subsection{Settings}
\label{sec:settings}
\textbf{Datasets:}

The ASVspoof2019 simulated PA dataset  \citep{DATASET}, an improved version with controlled environments  and acoustic configurations of the dataset of 2017, can be divided into three subsets: the training set (PA train set), the development set (PA dev set), and the evaluation set (PA eval set).
%  Attacks in the training and development sets are generated with the same simulation algorithm/conditions, while attacks in the evaluation set are additionally generated with different and unseen simulation algorithm/conditions to simulate unknown attacks. All these three subsets have no speaker overlap. The number of bonafide and spoofed utterances is about 218,000. 

%Evaluation data is about 8.5 GB generated by the same way as training and development data with different acoustic and replay configurations. 

In addition to the simulated PA subsets, we use the Real-PA dataset, recently released by the ASVspoof committee \citep{DATASET}, which {was} made with real replay operations from three different labs, to further test the 
{performance} of anti-spoofing systems. Note that the real data contains additive noise, which is not {contained} in the simulated PA subsets. As a consequence, spoofing detection results on the real replay data are not expected to be as good as those obtained from the simulated PA subsets, but we can still get some insights by comparing conventional methods with deep learning-based methods.

Details about the simulated PA datasets and the Real-PA dataset are illustrated in Table \ref{table:dataset} and  Table \ref{table:attackdefinition}.
\begin{table}[h]
\centering
\caption{Statistics of datasets used in this work}

%\resizebox{\textwidth}{!}{
\begin{tabular}{c|c|c}
\toprule[1.5pt]
Datasets & \# Bonafide utterances &  \# Spoofed utterances \\
\midrule[1pt]
PA train set & 5,400&48,600 \\
PA dev set &  5,400&24,300 \\
PA eval set & 18,090 &116,640 \\
Real-PA set & 540 &2,160 \\
\bottomrule[1.5pt]
\end{tabular}
%}
\label{table:dataset}
\end{table}

\begin{table}[h]
 \renewcommand\arraystretch{1.2}
 \centering

 \caption{Detailed information of different attack sources}

 \label{table:attackdefinition} 
\begin{tabular}{c|c|c|c}
\toprule[1.5pt]
\diagbox{Attack Factor}{Factor Type}
%\multirow{2}{*}[-0.5\dimexpr \aboverulesep + \belowrulesep + \cmidrulewidth]{Attack Factor}   & \multicolumn{3}{c}{Factor Type}  \\ 
%\cmidrule(lr){2-4}
&  A &B &C\\
\midrule[1pt]
Attacker-to-talker distance $D_a$ (cm)&$10-50$&$50-100$&$>100$\\
Replay device quality $Q$&perfect&high &low\\
\bottomrule[1.5pt]

\end{tabular}
 \end{table}

\begin{table*}[h]
\centering
\setlength{\belowcaptionskip}{10pt}
\caption{Overall Performance of different systems on the ASVspoof2019 PA eval set. We experimented with mean-fusion and logistic regression (LR) fusion for models that use three kinds of features all trained with {BCE or BFL}. The \textit{BFL + LR Fusion} achieves the best performance with only three single models. Better generalization ability of the proposed methods could be seen, i.e. reduced overfitting on the PA Dev Set.}

\resizebox{\textwidth}{!}{
  \begin{threeparttable}

\begin{tabular}{c|c|c||c|c|c|c}

\toprule[1.5pt]
\multirow{2}{*}[-0.5\dimexpr \aboverulesep + \belowrulesep + \cmidrulewidth]{Method} &    \multirow{2}{*}[-0.5\dimexpr \aboverulesep + \belowrulesep + \cmidrulewidth]{  System } & \multirow{2}{*}[-0.5\dimexpr \aboverulesep + \belowrulesep + \cmidrulewidth]{\shortstack{\# Models}}& \multicolumn{2}{c|}{PA Dev Set}  & \multicolumn{2}{c}{PA Eval Set}\\ 
\cmidrule(lr){4-5} \cmidrule(lr){6-7} 
& & &  t-DCF$_{norm}^{min}$ & EER(\%)& t-DCF$^{min}_{norm}$ & EER(\%)\\
\midrule[1pt]

\multirow{2}{*}[-0.5\dimexpr \aboverulesep + \belowrulesep + \cmidrulewidth]{Official Baseline}& LFCC+GMM & -\tnote{a} & 0.2554 & 11.96  &0.3017& 13.54\\
& CQCC+GMM& - & 0.1953 & 9.87  & 0.2454&11.04\\
\hline \hline

\multirow{1}{*}[-0.5\dimexpr \aboverulesep + \belowrulesep + \cmidrulewidth]{\cite{DKU2019}}& Fusion System &6 &0.0064&0.24& 0.0168&0.66\\
\hline \hline

\multirow{1}{*}[-0.5\dimexpr \aboverulesep + \belowrulesep + \cmidrulewidth]{\cite{ASSERT2019}}
& Fusion System &5 &\textbf{0.0030}&\textbf{0.13}& 0.0160&0.59\\
\hline \hline
\multirow{4}{*}[-0.5\dimexpr \aboverulesep + \belowrulesep + \cmidrulewidth]{D3M (This work)}& 

BCE + Mean Fusion  &3 &0.0092&0.40&0.0153&	0.62\\
&BCE + LR Fusion  &3 &0.0084&	0.37&0.0151&0.61\\
&BFL + Mean Fusion  &3 &0.0075&0.35&0.0127	&0.56\\
&BFL + LR Fusion  &3 &0.0077&	0.35&\textbf{0.0124}&	\textbf{0.55}\\
\bottomrule[1.5pt]
\end{tabular}
  \begin{tablenotes}
  \item[a] The official baseline adopts conventional methods and therefore does not participate in the comparison of the number of neural networks used for model ensemble.
  \end{tablenotes}
  \end{threeparttable}
}

\label{table:fusion}
\end{table*}

\textbf{Feature Extraction:} We use the following configurations to extract features:
\begin{itemize}
\item STFT-gram: STFT spectrum was extracted with 25-ms frame length and 10-ms frame shift. The number of FFT bins was set to 1,024. Spectrums of all the frames were then concatenated to form STFT-gram.

\item MGD-gram: Tuned on the development set, the parameters of the MGD function were empirically set to $\rho = 0.2$ and $\lambda = 0.7$. The number of FFT bins was 1,024. 

\item CQT-gram: CQT spectrum was extracted with a hop length of 128 sample points. The number of octaves and the number of bins per octave $b$ were set to 9 and 96, respectively.

\end{itemize}

\textbf{Training Scheme:} The networks were optimized by the AdamW optimizer \citep{adamw}, with  parameter settings $\beta_1 = 0.9$, $\beta_2 = 0.999$, and weight decay 5e-5, which can substantially improve the generalization performance of the widely-used optimizer Adam. Besides, scheduler \textit{ReduceLROnPlateau} in PyTorch was employed with max\_patience = 3 and reduce\_factor = 0.1 to reduce the learning rate once learning stagnate{d}.

\textbf{Evaluation Metrics:}
\begin{itemize}
\item Tandem detection cost function (t-DCF) : Introduced in the ASVspoof2019 Challenge, t-DCF reflects the influence of spoofing countermeasure (CM) performance on ASV reliability under the same analysis framework \citep{Kinnunen2018}. We adopt the t-DCF as the primary evaluation metric.

\item Equal Error Rate (EER): 
Determined by adjusting the threshold to make the false rejection rate (FRR) {equal} to false acceptance rate (FAR), EER is used as the secondary evaluation metric in our experiments, which is suitable for measuring the performance of a single anti-spoofing or ASV system.

\end{itemize}

\subsection{Evaluation Results on the ASVspoof2019 PA Eval Set}

\subsubsection{Overall Performance}

Table \ref{table:fusion} gives our overall performance on the PA eval set. We compare our models with the top-performing systems \cite{DKU2019} and \cite{ASSERT2019} ranked  3rd and 4th in the 2019 challenge.
Instead of these baseline models with five or more complex single models, our best fusion system outperforms them with only three single models, which is simple and computationally efficient. The results on the PA dev set and PA eval set also demonstrate the effectiveness of the balanced focal loss that focuses on harder samples to mitigate the discrepancy of data distributions between training and inference and make a model more generalizable.

\subsubsection{Results on MGD-gram vs GD-gram}
\begin{table}[h]
\centering
\setlength{\belowcaptionskip}{10pt}
\caption{Performance of MGD-gram vs GD-gram under different configurations on the ASVspoof2019 PA dev {set. No} other tricks like data augmentation were used in our experiments.}

%\resizebox{\textwidth}{!}{
\begin{tabular}{c|c||c|c}
\toprule[1.5pt]
Feature & Model &  t-DCF$^{min}_{norm}$ & EER(\%) \\
\midrule[1pt]
GD-gram & ResNet w BCE & 0.0467 & 1.81 \\
MGD-gram & ResNet w BCE & 0.0288 &	1.07\\
MGD-gram & ResNet w BFL & \textbf{0.0257} &	\textbf{1.04}\\
\bottomrule[1.5pt]
\end{tabular}
%}
\label{table:featurecompare}
\end{table}

%We first evaluate the performance of our proposed MGD-gram under the same condition (cross entropy loss) with GD-gram, and then compare the performance with loss functions.
The results in Table \ref{table:featurecompare} show that MGD-gram is significantly better than GD-gram (by 40.9\%) as a kind of feature representation.  Comparison in Table \ref{table:featurecompare} also verifies the effectiveness of the balanced focal loss. More detailed evaluation results for BFL will be reported in later subsections.

\subsubsection{Single Models}
%\# yhc and dyq
\label{analysisEVAL}
\begin{table}[h]
\centering
\setlength{\belowcaptionskip}{10pt}
\caption{Performance of different single models on the ASVspoof2019 PA eval set. EM refers to the expectation maximization algorithm for the GMM model training. }

\resizebox{\columnwidth}{!}{
\begin{tabular}{c|c|c||c|c|c|c}

\toprule[1.5pt]
\multirow{2}{*}[-0.5\dimexpr \aboverulesep + \belowrulesep + \cmidrulewidth]{Method} &    \multirow{2}{*}[-0.5\dimexpr \aboverulesep + \belowrulesep + \cmidrulewidth]{  Model } & \multirow{2}{*}[-0.5\dimexpr \aboverulesep + \belowrulesep + \cmidrulewidth]{\shortstack{Training\\Objective}}& \multicolumn{2}{c|}{PA Dev Set}  & \multicolumn{2}{c}{PA Eval Set}\\ 
\cmidrule(lr){4-5} \cmidrule(lr){6-7} 
& & &  t-DCF$_{norm}^{min}$ & EER(\%)& t-DCF$^{min}_{norm}$ & EER(\%)\\
\midrule[1pt]
% \multirow{2}{*}[-0.5\dimexpr \aboverulesep + \belowrulesep + \cmidrulewidth]{Official Baseline}& LFCC+GMM & EM & 0.2554 & 11.96  &0.3017& 13.54\\
% & CQCC+GMM& EM & 0.1953 & 9.87  & 0.2454&11.04\\
% \hline \hline
\multirow{1}{*}[-0.5\dimexpr \aboverulesep + \belowrulesep + \cmidrulewidth]{\cite{DKU2019}}& ResNet+GD-gram& BCE & 0.0467& 1.81 & 0.0439&1.79\\
\hline \hline

\multirow{1}{*}[-0.5\dimexpr \aboverulesep + \belowrulesep + \cmidrulewidth]{\cite{ASSERT2019}}& SENet+STFT-gram& BCE & 0.0150 & \textbf{0.58} & 0.036&1.29\\
\hline \hline
\multirow{6}{*}[-0.5\dimexpr \aboverulesep + \belowrulesep + \cmidrulewidth]{D3M  (This work)}& ResNet+STFT-gram& BCE & \textbf{0.0131}&	0.65&	0.0261&	1.12 \\
&\quad\quad\quad +STFT-gram& BFL &0.0163&	0.63&\textbf{0.0251}&\textbf{1.01}\\ 
&ResNet+MGD-gram& BCE & 0.0288&1.07&	0.0396&1.57\\ 
&\quad\quad\quad +MGD-gram& BFL & 0.0257&	1.04	&0.0343&1.39\\
 &ResNet+CQT-gram& BCE & 0.0445&	1.87&0.0477&	2.02\\ 
&\quad\quad\quad +CQT-gram& BFL & 0.0393	& 1.80	&0.0465&1.89\\ 

\bottomrule[1.5pt]
\end{tabular}
}
\label{table:single}
\end{table}

As shown in Table \ref{table:single}, on the PA eval set, for single models, all the deep learning-based models achieve better performance than the conventional GMM models.
In the ASVspoof2019 Challenge, \cite{DKU2019} proposes to model speech characteristics using the ResNet architecture \citep{RESNET} and phase-based GD-gram features. \cite{ASSERT2019}, using STFT-gram as the feature input, employs SENet which contains squeeze and excitation operations to facilitate the feature extraction and explicitly model the importance of different feature channels. Compared with them, our single models, with a similar architecture to \cite{DKU2019} and features introduced in Section \ref{sec:feature}, show 
competitive results. It is worth mentioning that the number of parameters in our models is less than that of \cite{ASSERT2019}. 
% We hope that less number of parameters and a simple architecture could make the evaluation convincing.

Comparing our models trained with BFL and BCE, {we find} that models equipped with BFL as the training objective significantly outperform those with BCE. Taking the\textit{ ResNet+MGD-gram} as an example, BFL improves the min-tDCF and EER by 12.6\% and 11.5\%, respectively. 
The \textit{ResNet+STFT-gram+BFL} model achieves the overall best performance and shows a better generalization ability in terms of the results on the PA dev set and {the} PA eval set when compared with \textit{SENet+STFT-gram} in \cite{ASSERT2019}. Detailed analysis for the effectiveness of the training objective with respect to each attack type will be reported in Subsection \ref{sec:analysisLOSS}.

%\subsubsection{Statistical Significance Test }
%Paired Student's t-test was performed here with null hypothesis being that there are no significant differences between the performance of system BFL + LR Fusion and system BCE + LR Fusion, and observed differences are purely due to random and not systematic factors. \# Not finished yet.
\subsection{Effectiveness of the Training Objective}
\label{sec:analysisLOSS}
In this subsection, we present detailed performance analysis for the training objective with the results of the single models and fusion systems over nine attacks. 

% \begin{figure}[h]
% \centering
% \includegraphics[width=.6\columnwidth]{images/focalanalysis1.png} 
% \caption{The score distributions of single models in this paper, tested on the ASVspoof2019 PA eval set. Best viewed in color.}
% \label{fig:ana1}
% \end{figure}

% Fig. \ref{fig:ana1} shows the score distributions of single models used in Subsection \ref{analysisEVAL}.
% Obvious differences can be seen for each of the three models using BFL and BCE. For the three using BFL, the scores are more concentrated near the zero point, and as a consequence of the introduction of the focusing factor $\gamma$ and the balancing factor $\alpha$ in the balanced focal loss, when it is near the zero point, the slope is larger than that of the three ones using BCE, meaning that for the three independent models using BFL, the difference in score is more obvious than that of the other three ones using BCE, when judging the bonafide and spoofed utterances.

\begin{figure}[h]
\centering

\setlength{\abovecaptionskip}{1.5em}
\includegraphics[width=\columnwidth]{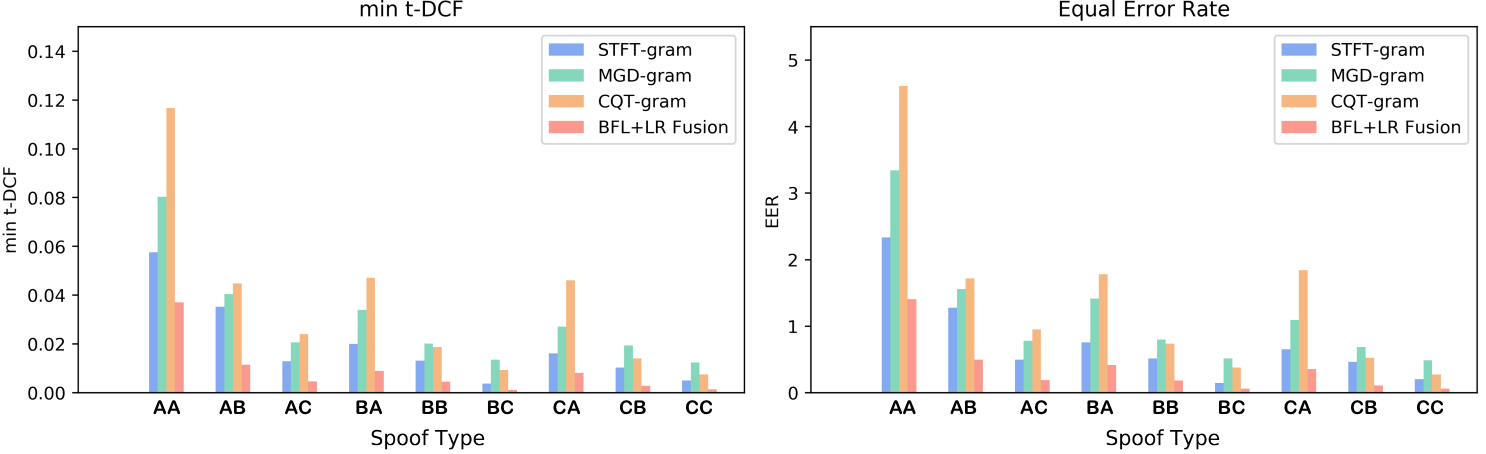} 
\caption{Detailed evaluation results over nine attack types for single models and the best fusion system \textit{BFL + LR Fusion} in terms of min-tDCF and EER. All the single models are trained with BFL. Best viewed in color.}
\label{fig:ana2}
\end{figure}

\begin{figure}[h]
\setlength{\abovecaptionskip}{1.5em}

\centering
\includegraphics[width=\columnwidth]{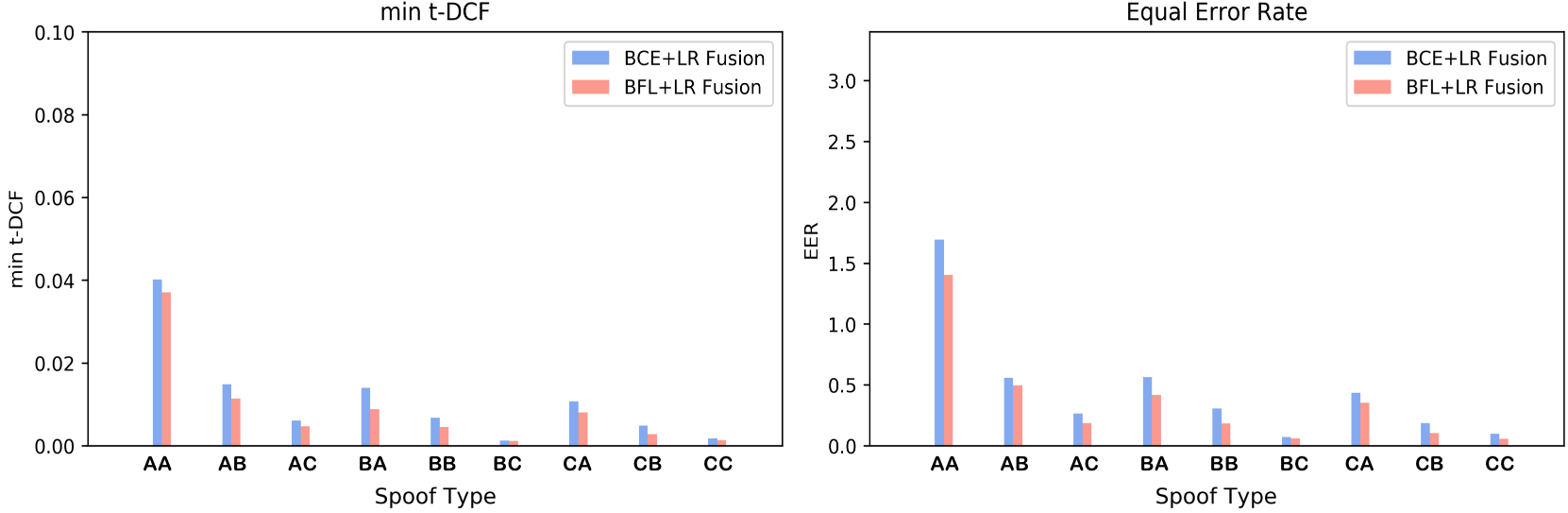} 

\caption{Comparision of fusion systems \textit{BCE+LR Fusion} and \textit{BFL+LR Fusion} based on the balanced focal loss and cross-entropy loss in terms of min-tDCF and EER over nine attack types. Best viewed in color.}
% 这里是想说the balanced (focal loss and cross entropy loss)，是不是不用在cross前面加the呢
\label{fig:ana3}
\end{figure}

To make detailed analysis for the effectiveness of the training objective, we divide the PA Eval Set into nine parts based on the types of spoofing attack information, corresponding to nine different spoofing methods: AA, AB, AC, BA, BB, BC, CA, CB, and CC. We experiment with single attacks and then plot the score statistics in Fig. \ref{fig:ana2} and Fig. \ref{fig:ana3}.
\vspace*{1em}

As shown in Fig.\ref{fig:ana2}, the three complementary independent models using BFL and the fusion system that combining these three have a good ability to discriminate against every type of spoof information, and for types that are more difficult and closer to the bonafide ones such as AB, BA, and CA,  those trained with BFL still have strong ability to distinguish the bonafide and spoofed utterances. In particular, for type AA, which is almost indistinguishable from bonafide utterances, the system using BFL also achieves lower min-tDCF and EER. 

Fig.\ref{fig:ana3} shows the performance of  two fusion systems with two different loss functions over nine types of spoofing attack methods. As can be seen from this figure, the use of BFL makes \textit{BFL+LR Fusion} perform much better than \textit{BCE+LR Fusion}. It is worth mentioning that for the quality attack samples of type AA, which is difficult to distinguish, the fusion system using BFL is more distinguishable than the fusion system using the widely-used balanced cross-entropy loss. This also reflects the robustness and generalization ability of the \textit{BFL+LR Fusion} system, dynamically focusing more on indistinguishable samples.

\subsection{Understanding the Network Decisions with Saliency Maps via \textcolor{black}{Backpropagation}}

\begin{figure*}[h]
\centering
\includegraphics[width=\textwidth]{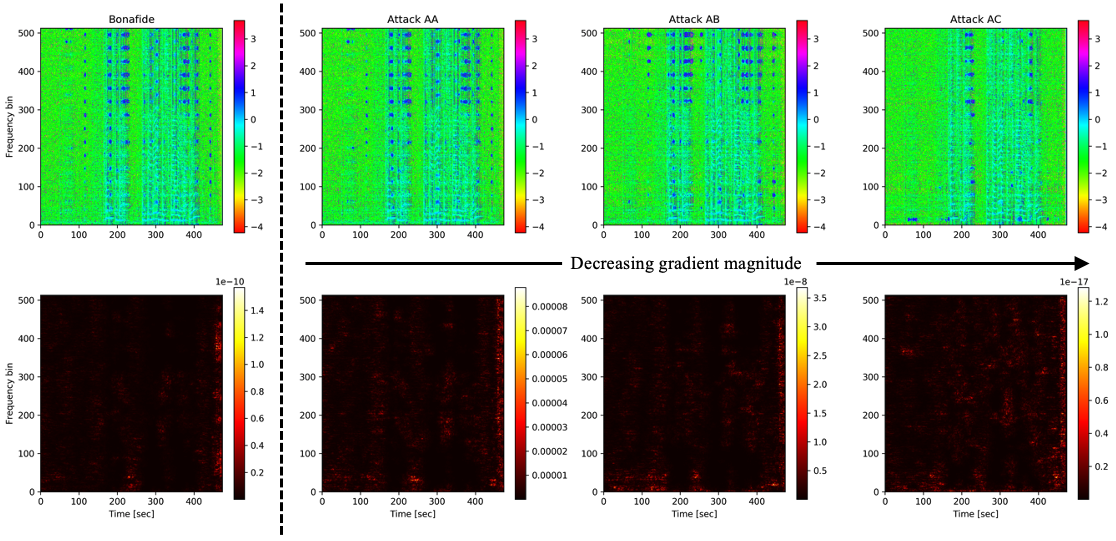} 
\caption{Visualization of original features (\textbf{Top}) vs. saliency maps (\textbf{Bottom}) via backpropagation to better understand critical parts for the network's making classification decisions. Hotness corresponds to regions that have a big impact on the final decisions. Decreasing gradient magnitude and increasing amount of hot pixels (time-frequency bins) verified the proposed idea, since higher gradient magnitude of harder attack types, with fewer amount of hot pixels, implies more decision sensitivity to value changes in corresponding pixels for utterances generated by quality attacks, hence these samples are harder.}
\label{fig:sali}
\end{figure*}

To better understand how the network make its classification decisions and verify our original motivation of leveraging BFL as the proper training objective to solve the data discrepancy problem, we further visualize the the original features and their corresponding saliency maps via \textcolor{black}{backpropagation} \cite{sali_vis}, an efficient way of network visualization. As shown in Fig. \ref{fig:sali}, the magnitudes of saliency maps are decreasing from quality attack type AA to relatively easy type AC. The decreasing gradient magnitude, together with increasing amount of hot ``pixels'' (time-frequency bins), verifies the initial intuition (see Fig.\ref{fig:main}) and demonstrates the necessity of our proposed method.

% \subsection{Evaluation Results on Different Model Architectures}
% To prove that the performance gain by BFL is model-independent, we replace our model backbone ResNet with the stronger SENet architecture \citep{ASSERT2019}. As presented in Table \ref{table:diffbackbone}, we find that model \textit{SENet w BFL} achieves the best performance, demonstrating SENet architecture's superior modeling capabilities, and the comparable results between \textit{SENet w BFL} and \textit{ResNet w BFL} imply the effectiveness of BFL training objective.

% \begin{table}[h]
% \centering
% %\setlength{\belowcaptionskip}{10pt}
% \caption{Performance of different backbone architectures on the ASVspoof2019 PA eval set (single model). MGD-gram feature is used for training. For SENet, we report best results with its reduction ratio parameter being $r=4$.}

% \begin{tabular}{c|c||c|c}
% \toprule[1.5pt]
%  \multirow{2}{*}[-0.5\dimexpr \aboverulesep + \belowrulesep + \cmidrulewidth]{  Model } & \multirow{2}{*}[-0.5\dimexpr \aboverulesep + \belowrulesep + \cmidrulewidth]{\shortstack{Training\\Objective}}& \multirow{2}{*}[-0.5\dimexpr \aboverulesep + \belowrulesep + \cmidrulewidth]{ t-DCF$^{min}_{norm}$} &
%  \multirow{2}{*}[-0.5\dimexpr \aboverulesep + \belowrulesep + \cmidrulewidth]{ EER(\%)}  \\
% &&& \\

% \midrule[1pt]
%  ResNet & BCE & 0.0396 & 1.57 \\
%  SENet &  BCE &  0.0349& 1.46\\
%  ResNet & BFL & 0.0343 & 1.39 \\
% SENet &  BFL & \textbf{0.0339} &\textbf{1.39}\\
% \bottomrule[1.5pt]
% \end{tabular}

% \label{table:diffbackbone}

% \end{table}

\subsection{Evaluation Results on the Real-PA Set}
%\# dyq and yhc

\begin{table}[h]
\centering
\caption{Performance of different systems on the ASVspoof2019 \textbf{Real-PA} set.  For fusion systems, only the result of the best (Mean-Fusion) is reported in this table.\\\\}

%\resizebox{\textwidth}{!}{
\begin{tabular}{c|c||c}

\toprule[1.5pt]
 \multirow{2}{*}[-0.5\dimexpr \aboverulesep + \belowrulesep + \cmidrulewidth]{  System } & \multirow{2}{*}[-0.5\dimexpr \aboverulesep + \belowrulesep + \cmidrulewidth]{\shortstack{Training\\Objective}}& \multirow{2}{*}[-0.5\dimexpr \aboverulesep + \belowrulesep + \cmidrulewidth]{ EER(\%)} \\
&&\\
\midrule[1pt]

LFCC+GMM & EM & 28.92 \\
 CQCC+GMM& EM & \textbf{12.40} \\
\hline \hline
ResNet+STFT-gram& BCE & 46.85\\
\quad\quad\quad +STFT-gram& BFL &51.50\\ 
ResNet+MGD-gram& BCE &23.91\\ 
\quad\quad\quad +MGD-gram& BFL&29.86\\
 ResNet+CQT-gram& BCE & 	30.95\\ 
\quad\quad\quad +CQT-gram& BFL &23.56\\ 
\hline
Mean-Fusion&BFL &25.02 \\
\bottomrule[1.5pt]
\end{tabular}
%}
\label{table:realres}

\end{table}
% \begin{figure}[h]
% \centering
% \includegraphics[width=.6\columnwidth]{images/realscoredistribution.png} 
% \caption{Score distributions of different systems, tested on the ASVspoof2019 Real-PA eval set. Best viewed in color.}
% \label{fig:real}
% \end{figure}

The performance of different systems on the ASVspoof2019 Real-PA set are shown in Table \ref{table:realres}. To our knowledge, we are the first to report the evaluation results on the recently-released Real-PA dataset \cite{DATASET}, and unexpected experimental results are shown which are worthy of discussion.
We can observe that current top-performing systems on real data are not as good as on the ASVspoof2019 simulated PA dataset. Although there is additional noise in the Real-PA set which is not contained in the  simulated PA dataset, conventional GMM models still perform best. The performance degradation may be due to the fact that simulated data cannot match real scenarios completely. Another possible reason may be the time-frequency distortions captured by CNN-based methods like ResNet are not detected by the time-level GMM model, which also needs further analysis and improvement. 
% ResNet as feature extractors for anti-spoofing still need further exploration and improvement.
\section{Conclusions}
%\# all of us
This paper aimed at resolving two challenges when designing replay attack detection systems. Firstly, we proposed D3M to leverage the novel balanced focal loss to dynamically mitigate the discrepancy of the data distributions between training and inference. We then presented the experiments with MGD-gram and selected complementary features on the ASVspoof2019 dataset.
Experimental results and detailed analysis verified the effectiveness of the proposed methods by comparing them with the top-performing systems from the 2019 Challenge.
% Our complementary fusion system with only three kinds of features yields an relative improvement of 22.5\% and 7\% for min-tDCF and EER, respectively, comparing with the top-performing systems containing five or more single models. 
% This demonstrates the superiority of the use and selection of complementary features. 

Moreover, we analyzed the unexpected performance of deep learning-based methods under real data. We hence argue that besides additive noise there may be other mismatch between real data (ASVspoof2019 Real-PA set) and simulated data (used in the 2019 Challenge), or that time-level GMM backends are more robust to time-frequency distortions than neural networks. In the future, we will dive more into bridge the huge gap for deep learning frameworks in real scenarios. We believe that integrating information produced by conventional models will be beneficial, which needs further explorations.

\section*{Acknowledgments}

The authors would like to thank the reviewers for their constructive and insightful suggestions, and the ASVspoof Committee for  preparing the dataset, and organizing the challenge. 
 
%\section*{References}
%\cite{IEEEexample:articleetal}
%\cite{Zhao2007Voltage}
%Please number citations consecutively within brackets \cite{b1}. The 
%sentence punctuation follows the bracket \cite{b2}. Refer simply to the reference 
%number, as in \cite{b3}---do not use ``Ref. \cite{b3}'' or ``reference \cite{b3}'' except at 
%the beginning of a sentence: ``Reference \cite{b3} was the first $\ldots$''
%
%Number footnotes separately in superscripts. Place the actual footnote at 
%the bottom of the column in which it was cited. Do not put footnotes in the 
%abstract or reference list. Use letters for table footnotes.
%
%Unless there are six authors or more give all authors' names; do not use 
%``et al.''. Papers that have not been published, even if they have been 
%submitted for publication, should be cited as ``unpublished'' \cite{b4}. Papers 
%that have been accepted for publication should be cited as ``in press'' \cite{b5}. 
%Capitalize only the first word in a paper title, except for proper nouns and 
%element symbols.
%
%For papers published in translation journals, please give the English 
%citation first, followed by the original foreign-language citation \cite{b6}.
\bibliographystyle{IEEEtran}
\bibliography{article}

% Generated by IEEEtran.bst, version: 1.12 (2007/01/11)
\begin{thebibliography}{10}
\providecommand{\url}[1]{#1}
\csname url@samestyle\endcsname
\providecommand{\newblock}{\relax}
\providecommand{\bibinfo}[2]{#2}
\providecommand{\BIBentrySTDinterwordspacing}{\spaceskip=0pt\relax}
\providecommand{\BIBentryALTinterwordstretchfactor}{4}
\providecommand{\BIBentryALTinterwordspacing}{\spaceskip=\fontdimen2\font plus
\BIBentryALTinterwordstretchfactor\fontdimen3\font minus
  \fontdimen4\font\relax}
\providecommand{\BIBforeignlanguage}[2]{{%
\expandafter\ifx\csname l@#1\endcsname\relax
\typeout{** WARNING: IEEEtran.bst: No hyphenation pattern has been}%
\typeout{** loaded for the language `#1'. Using the pattern for}%
\typeout{** the default language instead.}%
\else
\language=\csname l@#1\endcsname
\fi
#2}}
\providecommand{\BIBdecl}{\relax}
\BIBdecl

\bibitem{LCNN2017}
\BIBentryALTinterwordspacing
G.~Lavrentyeva, S.~Novoselov, E.~Malykh, A.~Kozlov, O.~Kudashev, and
  V.~Shchemelinin, ``Audio replay attack detection with deep learning
  frameworks,'' in \emph{Proc. Interspeech 2017}, 2017, pp. 82--86. [Online].
  Available: \url{http://dx.doi.org/10.21437/Interspeech.2017-360}
\BIBentrySTDinterwordspacing

\bibitem{Kinnunen2018}
T.~Kinnunen, K.~A. Lee, H.~Delgado, N.~Evans, M.~Todisco, M.~Sahidullah,
  J.~Yamagishi, and D.~A. Reynolds, ``t-dcf: a detection cost function for the
  tandem assessment of spoofing countermeasures and automatic speaker
  verification,'' in \emph{Odyssey}, 2018, pp. 312--319.

\bibitem{LCNN2019}
\BIBentryALTinterwordspacing
G.~Lavrentyeva, S.~Novoselov, A.~Tseren, M.~Volkova, A.~Gorlanov, and
  A.~Kozlov, ``{STC Antispoofing Systems for the ASVspoof2019 Challenge},'' in
  \emph{Proc. Interspeech 2019}, 2019, pp. 1033--1037. [Online]. Available:
  \url{http://dx.doi.org/10.21437/Interspeech.2019-1768}
\BIBentrySTDinterwordspacing

\bibitem{ASSERT2019}
\BIBentryALTinterwordspacing
C.-I. Lai, N.~Chen, J.~Villalba, and N.~Dehak, ``{ASSERT: Anti-Spoofing with
  Squeeze-Excitation and Residual Networks},'' in \emph{Proc. Interspeech
  2019}, 2019, pp. 1013--1017. [Online]. Available:
  \url{http://dx.doi.org/10.21437/Interspeech.2019-1794}
\BIBentrySTDinterwordspacing

\bibitem{Challenge2019results}
\BIBentryALTinterwordspacing
M.~Todisco, X.~Wang, V.~Vestman, M.~Sahidullah, H.~Delgado, A.~Nautsch,
  J.~Yamagishi, N.~Evans, T.~H. Kinnunen, and K.~A. Lee, ``Asvspoof 2019:
  Future horizons in spoofed and fake audio detection,'' in \emph{Proc.
  Interspeech 2019}, 2019, pp. 1008--1012. [Online]. Available:
  \url{http://dx.doi.org/10.21437/Interspeech.2019-2249}
\BIBentrySTDinterwordspacing

\bibitem{Alzantot2019}
\BIBentryALTinterwordspacing
M.~Alzantot, Z.~Wang, and M.~B. Srivastava, ``{Deep Residual Neural Networks
  for Audio Spoofing Detection},'' in \emph{Proc. Interspeech 2019}, 2019, pp.
  1078--1082. [Online]. Available:
  \url{http://dx.doi.org/10.21437/Interspeech.2019-3174}
\BIBentrySTDinterwordspacing

\bibitem{DKU2019}
\BIBentryALTinterwordspacing
W.~Cai, H.~Wu, D.~Cai, and M.~Li, ``{The DKU Replay Detection System for the
  ASVspoof 2019 Challenge: On Data Augmentation, Feature Representation,
  Classification, and Fusion},'' in \emph{Proc. Interspeech 2019}, 2019, pp.
  1023--1027. [Online]. Available:
  \url{http://dx.doi.org/10.21437/Interspeech.2019-1230}
\BIBentrySTDinterwordspacing

\bibitem{Cui2019}
Y.~Cui, M.~Jia, T.-Y. Lin, Y.~Song, and S.~Belongie, ``Class-balanced loss
  based on effective number of samples,'' in \emph{Computer Vision and Pattern
  Recognition (CVPR)}, Long Beach, CA, 2019.

\bibitem{Focalloss}
T.~{Lin}, P.~{Goyal}, R.~{Girshick}, K.~{He}, and P.~{Dollar}, ``Focal loss for
  dense object detection,'' \emph{IEEE Transactions on Pattern Analysis and
  Machine Intelligence}, pp. 1--1, 2018.

\bibitem{featuredecision}
\BIBentryALTinterwordspacing
S.~{M S} and H.~Murthy, ``Decision-level feature switching as a paradigm for
  replay attack detection,'' in \emph{Proc. Interspeech 2018}, 2018, pp.
  686--690. [Online]. Available:
  \url{http://dx.doi.org/10.21437/Interspeech.2018-1494}
\BIBentrySTDinterwordspacing

\bibitem{Tom2018}
\BIBentryALTinterwordspacing
F.~Tom, M.~Jain, and P.~Dey, ``End-to-end audio replay attack detection using
  deep convolutional networks with attention,'' in \emph{Proc. Interspeech
  2018}, 2018, pp. 681--685. [Online]. Available:
  \url{http://dx.doi.org/10.21437/Interspeech.2018-2279}
\BIBentrySTDinterwordspacing

\bibitem{mgd2004}
R.~M. {Hegde}, H.~A. {Murthy}, and G.~V.~R. {Rao}, ``Application of the
  modified group delay function to speaker identification and discrimination,''
  in \emph{2004 IEEE International Conference on Acoustics, Speech, and Signal
  Processing}, vol.~1, May 2004, pp. I--517.

\bibitem{DATASET}
\BIBentryALTinterwordspacing
J.~Yamagishi, M.~Todisco, M.~Sahidullah, H.~Delgado, X.~Wang, N.~Evans,
  T.~Kinnunen, K.~A. Lee, V.~Vestman, and A.~Nautsch.
  \BIBforeignlanguage{en}{Asvspoof 2019: The 3rd automatic speaker verification
  spoofing and countermeasures challenge database}. Asvspoof2019 evaluation
  plan.pdf. [Online]. Available: \url{https://doi.org/10.7488/ds/2555}
\BIBentrySTDinterwordspacing

\bibitem{mgdxiao2015a}
\BIBentryALTinterwordspacing
X.~Xiao, X.~Tian, S.~Du, H.~Xu, E.~S. Chng, and H.~Li, ``Spoofing speech
  detection using high dimensional magnitude and phase features: the ntu
  approach for asvspoof 2015 challenge,'' in \emph{INTERSPEECH-2015}, 2015, pp.
  2052--2056. [Online]. Available:
  \url{https://www.isca-speech.org/archive/interspeech_2015/i15_2052.html}
\BIBentrySTDinterwordspacing

\bibitem{mgdliuyi2015a}
Y.~Liu, Y.~Tian, L.~He, J.~Liu, and M.~T. Johnson, ``Simultaneous utilization
  of spectral magnitude and phase information to extract supervectors for
  speaker verification anti-spoofing,'' in \emph{INTERSPEECH-2015}, 2015, pp.
  2082--2086.

\bibitem{cqccpropose}
\BIBentryALTinterwordspacing
M.~{T}odisco, H.~{D}elgado, and N.~{E}vans, ``{C}onstant {Q} cepstral
  coefficients: {A} spoofing countermeasure for automatic {S}peaker
  verification,'' \emph{{C}omputer {S}peech \& {L}anguage, 20 {F}ebruary 2017},
  02 2017. [Online]. Available: \url{http://www.eurecom.fr/publication/5146}
\BIBentrySTDinterwordspacing

\bibitem{mgd2003}
H.~A. {Murthy} and V.~{Gadde}, ``The modified group delay function and its
  application to phoneme recognition,'' in \emph{2003 IEEE International
  Conference on Acoustics, Speech, and Signal Processing, 2003. Proceedings.
  (ICASSP '03).}, vol.~1, April 2003, pp. I--68.

\bibitem{Zhao2007Voltage}
\BIBentryALTinterwordspacing
F.~Z. Zhao and R.~G. Yang, ``Voltage sag disturbance detection based on short
  time fourier transform,'' \emph{Proceeding of the CSEE}, vol.~27, no.~10, pp.
  27--28, 2007. [Online]. Available:
  \url{http://www.pcsee.org/EN/abstract/abstract18630.shtml}
\BIBentrySTDinterwordspacing

\bibitem{Brown1998Calculation}
J.~C. Brown, ``Calculation of a constant q spectral transform,'' \emph{Journal
  of the Acoustical Society of America}, vol.~89, no.~1, pp. 425--434, 1998.

\bibitem{NEURIPS2019_9015}
\BIBentryALTinterwordspacing
``Pytorch: An imperative style, high-performance deep learning library,'' in
  \emph{Advances in Neural Information Processing Systems 32}, 2019, pp.
  8024--8035. [Online]. Available: \url{https://arxiv.org/abs/1912.01703}
\BIBentrySTDinterwordspacing

\bibitem{adamw}
\BIBentryALTinterwordspacing
I.~Loshchilov and F.~Hutter, ``Decoupled weight decay regularization,'' in
  \emph{International Conference on Learning Representations}, 2019. [Online].
  Available: \url{https://openreview.net/forum?id=Bkg6RiCqY7}
\BIBentrySTDinterwordspacing

\bibitem{RESNET}
\BIBentryALTinterwordspacing
K.~He, X.~Zhang, S.~Ren, and J.~Sun, ``Deep residual learning for image
  recognition,'' \emph{CoRR}, vol. abs/1512.03385, 2015. [Online]. Available:
  \url{http://arxiv.org/abs/1512.03385}
\BIBentrySTDinterwordspacing

\bibitem{sali_vis}
K.~Simonyan, A.~Vedaldi, and A.~Zisserman, ``Deep inside convolutional
  networks: Visualising image classification models and saliency maps.''
  \emph{CoRR}, vol. abs/1312.6034, 2013.

\end{thebibliography}

%\begin{thebibliography}{00}
%\bibitem{b1} G. Eason, B. Noble, and I. N. Sneddon, ``On certain integrals of Lipschitz-Hankel type involving products of Bessel functions,'' Phil. Trans. Roy. Soc. London, vol. A247, pp. 529--551, April 1955.
%\bibitem{b2} J. Clerk Maxwell, A Treatise on Electricity and Magnetism, 3rd ed., vol. 2. Oxford: Clarendon, 1892, pp.68--73.
%\bibitem{b3} I. S. Jacobs and C. P. Bean, ``Fine particles, thin films and exchange anisotropy,'' in Magnetism, vol. III, G. T. Rado and H. Suhl, Eds. New York: Academic, 1963, pp. 271--350.
%\bibitem{b4} K. Elissa, ``Title of paper if known,'' unpublished.
%\bibitem{b5} R. Nicole, ``Title of paper with only first word capitalized,'' J. Name Stand. Abbrev., in press.
%\bibitem{b6} Y. Yorozu, M. Hirano, K. Oka, and Y. Tagawa, ``Electron spectroscopy studies on magneto-optical media and plastic substrate interface,'' IEEE Transl. J. Magn. Japan, vol. 2, pp. 740--741, August 1987 [Digests 9th Annual Conf. Magnetics Japan, p. 301, 1982].
%\bibitem{b7} M. Young, The Technical Writer's Handbook. Mill Valley, CA: University Science, 1989.
%\end{thebibliography}

\end{document}